%% file: latex/acl_latex.tex
\documentclass[11pt]{article}

\usepackage[final]{acl}

\usepackage{times}
\usepackage{latexsym}
\usepackage{multirow}
\usepackage{amsmath}
\usepackage{amssymb}
\usepackage[ruled,vlined,linesnumbered]{algorithm2e}

\usepackage[T1]{fontenc}

\usepackage[utf8]{inputenc}

\usepackage{microtype}

\usepackage{inconsolata}

\usepackage{graphicx}

\newcommand{\name}{\textbf{Response-G1}}
\newcommand{\namewithspace}{\textbf{Response-G1 }}

\title{\name: Explicit Scene Graph Modeling for Proactive Streaming Video Understanding}

\author{
 \textbf{Ke Ma\textsuperscript{1,2$\dagger$}},
 \textbf{Jiaqi Tang\textsuperscript{3$\dagger$}},
 \textbf{Bin Guo\textsuperscript{1*}},
 \textbf{Xueting Han\textsuperscript{1}},
 \textbf{Ruonan Xu\textsuperscript{1}},
 \textbf{Qingfeng He\textsuperscript{2}},
\\
 \textbf{Ziheng Wang\textsuperscript{1}},
 \textbf{Xu Wang\textsuperscript{2}},
 \textbf{Qifeng Chen\textsuperscript{3}},
 \textbf{Zhiwen Yu\textsuperscript{1,4}},
 \textbf{Yunhao Liu\textsuperscript{2*}}
\\
 \textsuperscript{1}Northwestern Polytechnical University,
 \textsuperscript{2}Tsinghua University
\\
 \textsuperscript{3}The Hong Kong University of Science and Technology,
 \textsuperscript{4}Harbin Engineering University
\\
 \small{
   \textbf{Project Page:} \url{https://github.com/kadmkbl/Response-G1}
 }
}

\begin{document}
\maketitle
\footnotetext{$\dagger$ These authors contributed equally to this work.}
\footnotetext{* Corresponding authors. \tt guob@nwpu.edu.cn, yunhao@tsinghua.edu.cn}
\input{sec/0_abstract}
\input{sec/1_intro}

\input{sec/2_related}
\input{sec/3_method}
\input{sec/4_experiment}
\input{sec/5_conclusion}

\newpage
\section*{Limitations}
\label{sec:limitation}
First, explicit scene graph modeling improves evidence-condition alignment, but its object-relation representation and similarity-based retrieval do not fully address all reasoning needs (e.g., "why"-style questions), where benefits may be limited. Future work could explore richer structural formulations (e.g., causal relations) and more memory and retrieval designs beyond top-K matching.
Second, the fixed clip size for online scene graph generation could be improved by incorporating event-level or semantics-level perception. Adaptive mechanisms that dynamically determine when to trigger generation and how many frames to include would enhance both efficiency and temporal modeling.
Third, LLM-based open-set scene graph generation avoids predefined vocabularies but faces a trade-off between query relevance and hallucination. While we rely on fine-tuning-free prompting, task-specific fine-tuning of the generator could yield more relevance-factuality balancing.

\section*{Acknowledgments}
This work was supported by the National Natural Science Foundation of China (No. U25B2042, 62532009, 62232004, 62332016, 62302259), and the Research Grants Council of HKSAR under grant number AoE/E-601/24-N.

\bibliography{custom}


\newpage
\input{sec/appendix}


\end{document}

%% file: sec/0_abstract.tex
\begin{abstract}

Proactive streaming video understanding requires Video‑LLMs to decide when to respond as a video unfolds, a task where existing methods often fall short due to their implicit, query‑agnostic modeling of visual evidence. 
We introduce \name, a novel framework that establishes explicit, structured alignment between the accumulated video evidence and the query’s expected response conditions via scene graphs. 
The framework operates in three fine‑tuning‑free stages: (1) online query‑guided scene graph generation from streaming clips; (2) memory‑based retrieval of the most semantically relevant historical scene graphs; and (3) retrieval‑augmented trigger prompting for per‑frame "silence/response" decisions. 
By grounding both evidence and conditions in a shared graph representation, \name\ achieves more interpretable and accurate response timing decisions. Experimental results on established benchmarks demonstrate the superiority of our method in both proactive and reactive tasks, validating the advantage of explicit scene graph modeling and retrieval in streaming video understanding.

\end{abstract}

%% file: sec/1_intro.tex
\begin{figure}[t]
\includegraphics[width=\columnwidth]{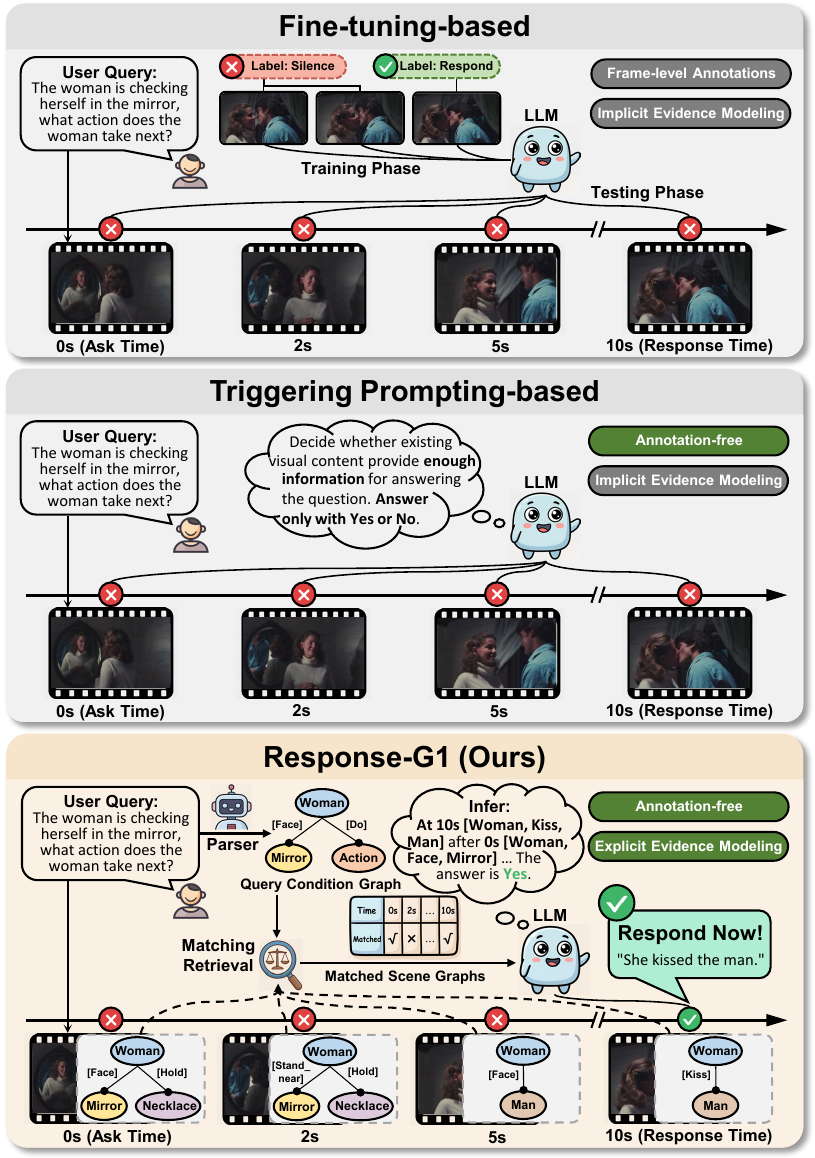}
\caption{Existing proactive mechanisms in streaming video understanding. By explicitly modeling both observed visual evidence and query-specific conditions via scene graphs, \namewithspace achieves accurate decisions of response timing.}
\label{fig:problem}
\end{figure}

\section{Introduction}
\label{sec:intro}

Video Large Language Models (Video-LLMs) have established themselves as a dominant paradigm for a wide range of video comprehension tasks~\cite{wang2025cascade}. A critical frontier within this domain is Streaming Video Understanding (SVU), which focuses on processing continuous, real-time video feeds~\cite{qian2024streaming, zhang2025flash}. Unlike offline analysis, SVU models must perceive, reason, and interact based solely on the incrementally observed visual content up to the current moment, making it indispensable for applications demanding immediate responsiveness, such as conversational AI assistants~\cite{wen2025ai} and autonomous embodied agents~\cite{zhang2025eyes}.

The prevalent paradigm in current SVU research is \textit{reactive interaction}, where models are designed to respond immediately to user queries~\cite{di2025streaming, xiong2025streaming, zeng2025streamforest}. However, this paradigm is fundamentally limited for queries that are predictive or anticipatory in nature, as the information required for a sufficient answer may only appear in future video segments. To address this, Video-LLMs need the capability for \textit{proactive interaction}, to autonomously determine the optimal moment to answer based on accumulated evidence~\cite{niu2025ovo}.

To implement proactive mechanisms, existing works have explored various approaches. One line of research~\cite{chen2024videollm, li2025lion} employs a streaming End-Of-Sequence (EOS) prediction pipeline, where the Video-LLM is trained to decode the EOS token to remain silent on frames that are not annotated for a response. Similarly relying on fine-tuning, another line of work~\cite{qian2025dispider, wang2025streambridge} introduces an auxiliary activation LLM with a binary classification head, which directly outputs a silence-or-response decision at each frame. However, as shown in Figure~\ref{fig:problem}, fine-tuning-based methods critically rely on detailed, frame-wise annotations, where consecutive and visually similar frames are often assigned opposite silence-or-response labels. This inconsistency significantly hinders the learning of a reliable decision boundary. Among fine-tuning-free approaches,~\cite{yao2025timechat} employs thresholds triggered by inter-frame differences. While simple, this method completely ignores query semantics, leading to suboptimal response timing.~\cite{yang2025streamagent} adopts a multi-LLM-agent framework, where response timing in video streams is determined through prompting a dedicated planner.

A fundamental limitation common to these approaches is their \textbf{implicit modeling} of the visual evidence and the response conditions implied by the query.
This implicit modeling limits the Video-LLM's ability to comprehensively understand and align the accumulated streaming evidence with query-specific conditions, thereby limiting the accuracy of response timing.
To overcome this, we argue that explicit representations are necessary, as they disentangle the underlying semantics and support more principled reasoning over streaming evidence.
Specifically, we observe that the user query typically depicts a scene that includes the objects and relations anticipated by the response conditions (see Figure~\ref{fig:problem}, which shows a query targeting a woman checking herself in the mirror).
Given that query-relevant scenes are inherently structured through objects and relations, we propose using scene graphs~\cite{johnson2015image} as a unifying representation to explicitly model both the visual evidence and the response conditions.

Driven by this insight, we propose \name, a novel framework that establishes a complete scene-graph-driven pipeline for proactive streaming video understanding.
Its core consists of three integrated components: (1) \textbf{Online Query-guided Scene Graph Generation}: We leverage Video-LLMs to abstract a scene graph for streaming video clips, focusing on salient, query-related evidence. (2) \textbf{Memory-based Scene Graph Retrieval}: A dynamic memory bank stores historical scene graphs. During streaming processing, we retrieve scene graphs most semantically relevant to the response conditions, providing a concise, aligned evidence context for decision-making. (3) \textbf{Retrieval-augmented Streaming Decision \& Response}: The retrieved scene graphs, interleaved with temporal cues, are fed into the Video-LLM alongside the visual tokens. A trigger-prompting mechanism enables per-frame silence-or-response decisions, and upon triggering, the final answer is generated using the same enriched context. Our framework operates in a fine-tuning-free manner, enhancing the model's inherent capabilities.

Extensive evaluations on StreamingBench~\cite{lin2024streamingbench} and OVO-Bench~\cite{niu2025ovo} demonstrate that \namewithspace significantly improves the accuracy of response timing and the quality of final answers, achieving state-of-the-art performance on streaming video understanding.

Our main contributions are:
\begin{itemize}
    \item We address proactive interaction in SVU through the explicit modeling of observed visual evidence and query-specific response conditions using structured scene graphs, offering a novel and interpretable pathway.
    \item  We design \namewithspace, an end-to-end, fine-tuning-free framework that seamlessly integrates online scene graph generation, memory-based graph retrieval, and retrieval-augmented streaming decision-making. Our framework, through explicit scene graph modeling and evidence retrieval, yields more accurate response timing decisions.
    \item Experimental results on established benchmarks show that our method sets a new state-of-the-art for both \textit{proactive} and \textit{reactive} streaming video understanding tasks.
\end{itemize}

%% file: sec/2_related.tex
\section{Related Works}
\label{sec:related}

\paragraph{Streaming Video Understanding.}
Streaming Video Understanding (SVU) refers to analyzing continuously arriving video streams without observing the complete video~\cite{zhou2024streaming, zheng2025hierarchical, wang2025enabling}.
Recently, Video-LLMs have achieved state-of-the-art performance on these tasks~\cite{tang2024hawk, bai2025qwen3vltechnicalreport}.
This line of research focuses primarily on modeling dynamic long-term context~\cite{huang2025online} and enabling real-time processing~\cite{zhang2025flash} within a reactive interaction framework, where streaming Video-LLMs must generate responses immediately as users pose queries at arbitrary timestamps during the video stream.

In practice, however, users may often issue queries that require future observations. This necessitates streaming Video-LLMs capable of proactive interaction, which must autonomously decide when the currently observed evidence satisfies the condition to respond, or otherwise remain silent.
To achieve query-aware proactive interaction, fine-tuning-based approaches train frame-level decision modules via streaming EOS token prediction~\cite{chen2024videollm, li2025lion} or auxiliary activation models~\cite{qian2025dispider, wang2025streambridge}, while fine-tuning-free methods primarily rely on prompting to query the Video-LLMs whether to remain silent at each frame~\cite{niu2025ovo, yang2025streamagent}.

However, such implicit modeling of visual evidence and response conditions hinders the understanding of response timing. In contrast, we introduce explicit scene graph modeling of both accumulated streaming evidence and query-specific conditions.
By feeding the aligned scene graph evidence into the Video-LLM, our method achieves more accurate response timing decisions, within a fine-tuning-free framework.

\paragraph{Scene Graph for Retrieval and Reasoning.}
Scene graphs provide a structured semantic representation of objects, attributes, and their relations, widely used for visual retrieval~\cite{johnson2015image} and spatio-temporal reasoning~\cite{xiao2023contrastive, chu2025fine}.
In retrieval, they enable accurate semantic-level similarity measurement between queries and visual databases~\cite{schroeder2020structured, yoon2021image}.
For proactive SVU tasks, the user query typically depicts an anticipated scene that includes target objects and relations.
Such a scene naturally lends itself to a structured graph representation.
Motivated by this, we propose a streaming pipeline that combines online scene graph generation with graph-retrieval augmentation. This allows us to explicitly model and reason over both the visual evidence and the response conditions, toward more accurate response timing decisions.

\paragraph{Scene Graph Generation.}
Scene Graph Generation (SGG) aims to parse visual inputs into structured object-relation graphs.
Traditional methods~\cite{cong2021spatial, nguyen2024hig} depend on closed-set detectors (e.g., Faster R-CNN~\cite{ren2015faster}),
limiting open-world applicability.
To address this, recent approaches leverage LLMs for open-vocabulary SGG~\cite{li2024pixels, yang2025llm, nguyen2025hyperglm}.
In our proactive SVU setting, we prompt the Video-LLM itself to generate scene graphs dynamically from the video stream, focusing on salient, query-related evidence.

%% file: sec/3_method.tex
\begin{figure*}[t]
\centering
  \includegraphics[width=1\textwidth]{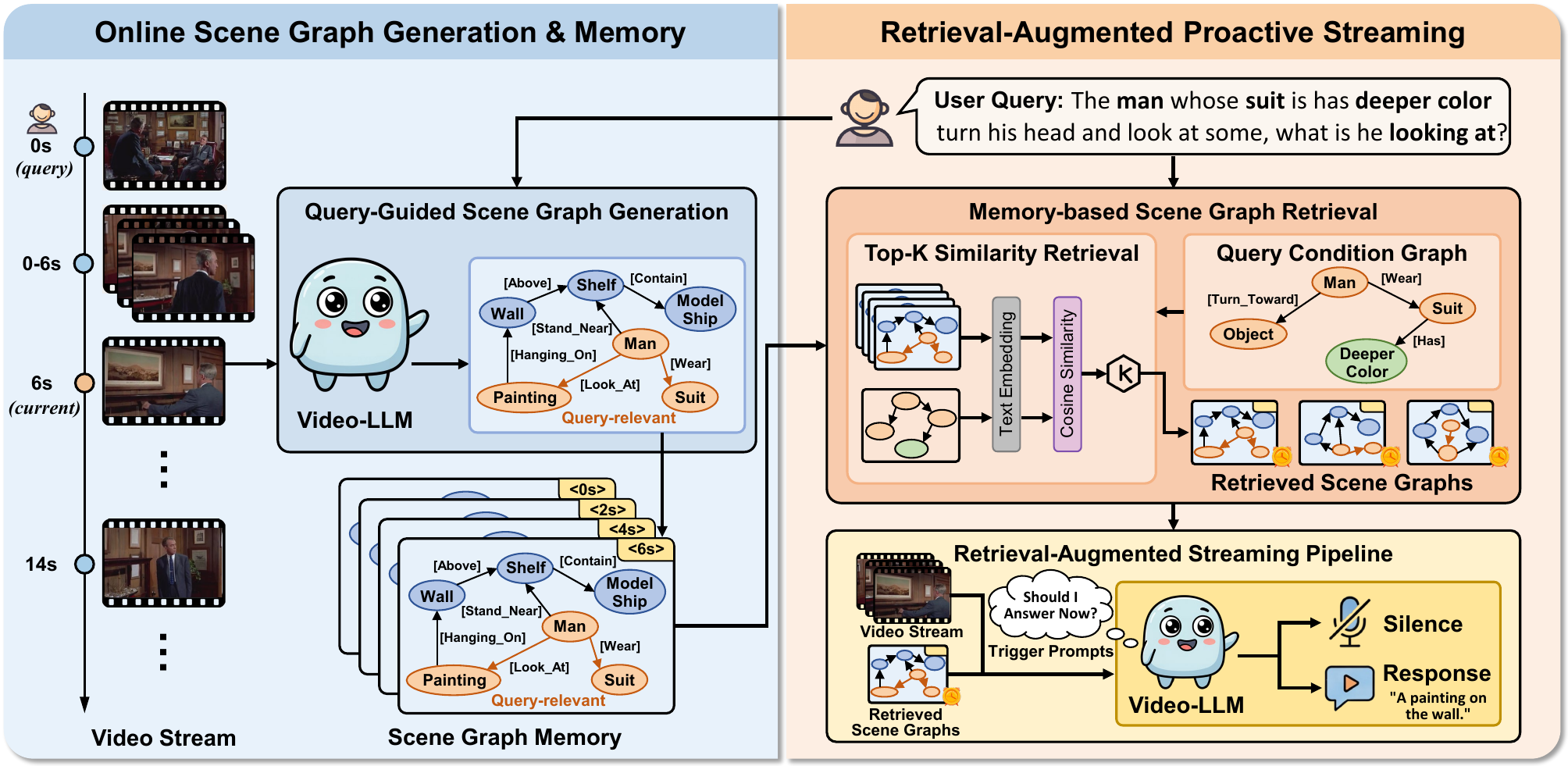}
  \caption{\textbf{Overview of the \namewithspace framework.} The system processes streaming video through three core components: (1) Online Query-Guided Scene Graph Generation, (2) Memory-Based Scene Graph Retrieval, and (3) Retrieval-Augmented Streaming Pipeline for proactive decision-making.}
  \label{fig:framework}
\end{figure*}

\section{Methodology}
\label{sec:method}

We propose \namewithspace, a novel framework that empowers Video-LLMs with proactive interaction capabilities by explicitly modeling the accumulated evidence and the response conditions through structured scene graphs. As illustrated in Figure~\ref{fig:framework}, our approach comprises three core components: (1) \textbf{online query-guided scene graph generation}, (2) \textbf{memory-based scene graph retrieval}, and (3) a \textbf{retrieval-augmented streaming pipeline}. We begin by formalizing the proactive streaming video understanding problem.

\subsection{Problem Formulation}
\label{subsec:problem}

Let $\mathcal{V}_T$ denote a streaming video of duration $T$, represented as a temporal sequence of sampled frames $\mathcal{F} = \{f_1, f_2, \dots, f_T\}$. At an arbitrary timestamp $t_{\text{ask}} \in [1, T]$ during the stream, a user issues a query $\mathcal{Q}_{t_{\text{ask}}}$.

\paragraph{Reactive vs. Proactive Paradigms.}
In the conventional reactive interaction paradigm, the Video-LLM is constrained to produce an immediate response at the query arrival moment. Formally, the response time $t_{\text{res}}$ equals $t_{\text{ask}}$, and the answer is generated based solely on the observed prefix $\mathcal{F}_{1:t_{\text{res}}}$.

In contrast, the proactive interaction paradigm allows the model to strategically delay its response until sufficient evidence is accumulated. The response time $t_{\text{res}}$ is a latent variable satisfying $t_{\text{ask}} \leq t_{\text{res}} \leq T$, determined dynamically by a per-frame decision function $\mathcal{D}(\cdot)$. At each time step $t \in [t_{\text{ask}}, t_{\text{res}}]$, the model evaluates whether the observed evidence in $\mathcal{F}_{1:t}$ satisfies the response conditions implicit in $\mathcal{Q}_{t_{\text{ask}}}$, outputting a proactive action $r_t \in \mathcal{R} = \{\texttt{silence}, \texttt{response}\}$.

\paragraph{Evidence-Condition Modeling via Scene Graphs.}
To achieve proactive response, we explicitly model both evidence and response conditions using structured scene graphs. Let $\mathcal{J}: (\mathcal{Q}, \mathcal{F}_{1:t}) \mapsto \mathbf{H}_t$ be a joint encoding function that maps the query and video frames into a unified representation space, where $\mathbf{H}_t \in \mathbb{R}^{d_h}$ denotes the joint hidden state. Let $\mathcal{S}: \mathbf{H}_t \mapsto \mathcal{G}_t$ be a scene graph extraction function that produces a structured graph $\mathcal{G}_t$ capturing objects, attributes, and their relations. Our proactive decision function is then formulated as:
\begin{equation}
\mathcal{D}\bigl(\mathbf{H}_t, \mathcal{S}(\mathbf{H}_t)\bigr) \rightarrow r_t.
\label{eq:proactive_decision}
\end{equation}
This explicit modeling enables more principled and interpretable reasoning over evidence-condition alignment, leading to more accurate response timing decisions.

\subsection{Online Query-Guided Scene Graph Generation}
\label{subsec:onlinesgg}

To explicitly model the evolving visual evidence, we design a one-stage, LLM-based framework for online scene graph generation from streaming video clips.

\paragraph{Formal Scene Graph Representation.}
For a video clip $\mathcal{C}_t$ centered at timestamp $t$, we generate a scene graph $\mathcal{G}_t = (\mathcal{O}_t, \mathcal{P}_t)$ via a prompted Video-LLM. Here, $\mathcal{O}_t$ is the set of nodes representing visual objects (e.g., person, car) and their attributes (e.g., red, large). $\mathcal{P}_t$ is the set of predicates (edges) capturing spatio-temporal relations (e.g., next\_to, holdin) between node pairs. Therefore, the graph can be represented as a collection of object-predicate-object triplets:
\begin{equation}
\mathcal{G}_t = \{\tau^{ij}_t = (o^i_t, p^{ij}_t, o^j_t) \mid o^i_t, o^j_t \in \mathcal{O}_t; p^{ij}_t \in \mathcal{P}_t\}.
\end{equation}

\paragraph{Query-Guided Generation for Relevance.}
To suppress irrelevant visual details and focus on query-salient evidence, we condition the scene graph generation on the user query $\mathcal{Q}$. Specifically, we inject $\mathcal{Q}$ into the generation prompt (see Appendix), steering the Video-LLM to prioritize query-relevant triplets. The query-guided generation process is formalized as:
\begin{equation}
\mathcal{G}_t = \mathcal{S}(\mathcal{C}_t; \mathcal{Q}).
\end{equation}
This directed generation enhances the relevance of the extracted scene graphs for subsequent evidence-condition matching.

\subsection{Memory-Based Scene Graph Retrieval}
\label{subsec:memorysgr}

To achieve fine-grained evidence-condition alignment, we introduce a memory module that stores historical scene graphs and retrieves the most query-relevant ones for semantic matching.

\paragraph{Textual Linearization and Embedding.}
For efficient retrieval, we linearize each scene graph triplet $\tau^{ij}_t$ into a natural language phrase $\phi^{ij}_t$ (e.g., $(woman, in, red) \rightarrow$ \textit{"woman in red"}). The full graph $\mathcal{G}_t$ is then represented as the concatenation of all its triplet phrases:
\begin{equation}
\Phi_t = \bigoplus_{i,j} \phi^{ij}_t.
\end{equation}
Similarly, we parse the user query $\mathcal{Q}$ into a query condition graph $\mathcal{G}_q$ (via the same Video-LLM) and derive its textual representation $\Phi_q$. Using $\Phi_q$ ensures format consistency with $\Phi_t$ for fair similarity computation.

We employ the Video-LLM's text encoder $\mathcal{E}_{\text{text}}(\cdot)$ to obtain dense embeddings. Let $\mathbf{E}_t = \mathcal{E}_{\text{text}}(\Phi_t) \in \mathbb{R}^{n_t \times d}$ and $\mathbf{E}_q = \mathcal{E}_{\text{text}}(\Phi_q) \in \mathbb{R}^{n_q \times d}$, where $n_t, n_q$ are token counts and $d$ is the embedding dimension. The graph representation is obtained via mean pooling over the token dimension:
\begin{align}
\mathbf{g}_t &= \text{MeanPool}(\mathbf{E}_t) \in \mathbb{R}^d, \\
\mathbf{g}_q &= \text{MeanPool}(\mathbf{E}_q) \in \mathbb{R}^d.
\end{align}

\paragraph{Similarity-Based Top-$K$ Retrieval.}
The semantic relevance between a clip-wise scene graph $\mathcal{G}_t$ and the query $\mathcal{Q}$ is quantified by cosine similarity:
\begin{equation}
\text{sim}(\mathcal{G}_t, \mathcal{Q}) = \frac{\mathbf{g}_t \cdot \mathbf{g}_q}{\|\mathbf{g}_t\| \|\mathbf{g}_q\|}.
\end{equation}
At each time step $t$, we maintain a memory bank $\mathcal{M}_t = \{\mathcal{G}_1, \dots, \mathcal{G}_t\}$ of generated scene graphs up to $t$. The top-$K$ most relevant graphs are retrieved:
\begin{equation}
\mathcal{G}^{\text{ctx}}_t = \{\mathcal{G}_{\tau} \mid \tau \in \text{TopK}\bigl(\{\text{sim}(\mathcal{G}_i, \mathcal{Q})\}_{i=1}^{t}, K\bigr)\},
\end{equation}
where $\mathcal{G}^{\text{ctx}}_t$ serves as the structured, response condition-aligned evidence context for the subsequent decision pipeline.

\subsection{Retrieval-Augmented Streaming Pipeline}
\label{subsec:pipeline}

The final component integrates the retrieved scene graphs into a streaming pipeline that performs per-frame decision-making and response generation.

\paragraph{Trigger Phase: Silence-or-Response Decision.}
At each time step $t \geq t_{\text{ask}}$, the model decides whether to respond. The input to the Video-LLM is constructed as a token sequence:
\begin{equation}
[\mathbf{f}_1, \dots, \mathbf{f}_t] \; \oplus \; \Psi(\mathcal{G}^{\text{ctx}}_t) \; \oplus \; \mathbf{p}_{\text{trg}},
\end{equation}
where $\mathbf{f}_i$ denotes the frame embedding for $f_i$, $\mathbf{p}_{\text{trg}}$ is the embedding of a trigger instruction (e.g., \textit{"Should I answer now? Yes or No."}), and $\Psi(\cdot)$ encodes the retrieved scene graphs with timestamps.

\paragraph{Timestamp-Aware Scene Graph Encoding.}
To provide temporal context, each retrieved graph $\mathcal{G}_i \in \mathcal{G}^{\text{ctx}}_t$ is prefixed with a textual timestamp token (e.g., \texttt{<2.0s>}) before encoding. Formally,
\begin{equation}
\Psi(\mathcal{G}^{\text{ctx}}_t) = \bigoplus_{i \in \mathcal{I}^{\text{ctx}}} \mathcal{E}_{\text{text}}\bigl(\texttt{<}t_i\texttt{s>} \oplus \Phi_i\bigr),
\end{equation}
where $\mathcal{I}^{\text{ctx}}$ is the set of retrieved indices, $t_i$ is the timestamp of $\mathcal{G}_i$, and $\Phi_i$ is its textual linearization. This encoding enhances the model's temporal reasoning about the evidence.

The Video-LLM processes this sequence and generates an interaction decision token (e.g., \texttt{Yes}/\texttt{No}). If the decision is \texttt{silence}, the process continues with the next frame. If \texttt{response}, the model proceeds to the response phase at $t_{\text{res}} = t$.

\paragraph{Response Phase: Answer Generation.}
Upon triggering, the final answer is generated using the context up to $t_{\text{res}}$. The input sequence is:
\begin{equation}
\label{eq:answer_generation}
[\mathbf{f}_1, \dots, \mathbf{f}_{t_{\text{res}}}] \; \oplus \; \Psi(\mathcal{G}^{\text{ctx}}_{t_{\text{res}}}) \; \oplus \; \mathbf{q},
\end{equation}
where $\mathbf{q} = \mathcal{E}_{\text{text}}(\mathcal{Q})$ is the embedding of the original user query. The Video-LLM then generates the natural language response.

\paragraph{Extension to Reactive Interaction.}
The same scene graph augmentation benefits traditional reactive interaction ($t_{\text{res}} = t_{\text{ask}}$). Using Equation~\ref{eq:answer_generation} with $\mathcal{G}^{\text{ctx}}_{t_{\text{ask}}}$, the model also achieves enhanced spatio-temporal grounding and answer quality through scene graph modeling, demonstrating the framework's versatility.

\begin{table*}[t]
\centering
\resizebox{1.0\linewidth}{!}{
\begin{tabular}{lccccccccccccccccc}
\hline
\multirow{2}{*}{\textbf{Model}} & \multicolumn{1}{c|}{\multirow{2}{*}{\textbf{Params}}} & \multicolumn{7}{c|}{\textbf{Real-Time Visual Perception}}                                                                    & \multicolumn{4}{c|}{\textbf{Backward Tracing}}                                  & \multicolumn{4}{c|}{\textbf{Forward Active Responding}}                         & \multirow{2}{*}{\textbf{\begin{tabular}[c]{@{}c@{}}Overall\\ Avg.\end{tabular}}}  \\ \cline{3-17}
                                & \multicolumn{1}{c|}{}                                 & \textbf{OCR} & \textbf{ACR} & \textbf{ATR} & \textbf{STU} & \textbf{FPD} & \textbf{OJR} & \multicolumn{1}{c|}{\textbf{Avg.}} & \textbf{EPM} & \textbf{ASI} & \textbf{HLD} & \multicolumn{1}{c|}{\textbf{Avg.}} & \textbf{REC} & \textbf{SSR} & \textbf{CRR} & \multicolumn{1}{c|}{\textbf{Avg.}} &                                   \\ \hline
\multicolumn{18}{c}{\textbf{Human}}                                                                                                                                                                                                                                                                                                                                                                                            \\ \hline
Human                           & \multicolumn{1}{c|}{-}                                & 94.0         & 92.6         & 94.8         & 92.7         & 91.1         & 94.0         & \multicolumn{1}{c|}{93.2}          & 92.6         & 93.0         & 91.4         & \multicolumn{1}{c|}{92.3}          & 95.5         & 89.7         & 93.6         & \multicolumn{1}{c|}{92.9}          & 92.8                              \\ \hline
\multicolumn{18}{c}{\textbf{Proprietary MLLMs}}                                                                                                                                                                                                                                                                                                                                                                                \\ \hline
GPT-4o~\cite{hurst2024gpt}                          & \multicolumn{1}{c|}{-}                                & 69.1         & 65.1         & 65.5         & 50.0         & 68.3         & 63.7         & \multicolumn{1}{c|}{63.6}          & 49.8         & 71.0         & 55.4         & \multicolumn{1}{c|}{58.7}          & 27.6         & 73.2         & 59.4         & \multicolumn{1}{c|}{53.4}          & 58.6                              \\
Gemini 1.5 Pro~\cite{team2024gemini}                  & \multicolumn{1}{c|}{-}                                & 87.3         & 67.0         & 80.2         & 54.5         & 68.3         & 67.4         & \multicolumn{1}{c|}{70.8}          & 68.6         & 75.7         & 52.7         & \multicolumn{1}{c|}{62.3}          & 35.5         & 74.2         & 61.7         & \multicolumn{1}{c|}{57.2}          & 65.3                              \\ \hline
\multicolumn{18}{c}{\textbf{Open-Source   Video-LLMs}}                                                                                                                                                                                                                                                                                                                                                                         \\ \hline
LongVU~\cite{shen2024longvu}                          & \multicolumn{1}{c|}{7B}                               & 55.7         & 49.5         & 59.5         & 48.3         & 68.3         & 63.0         & \multicolumn{1}{c|}{57.4}          & 43.1         & 66.2         & 9.1          & \multicolumn{1}{c|}{39.5}          & 16.6         & 69.0         & 60.0         & \multicolumn{1}{c|}{48.5}          & 48.5                              \\
InternVL-V2~\cite{chen2024far}                     & \multicolumn{1}{c|}{8B}                               & 68.5         & 58.7         & 69.0         & 44.9         & 67.3         & 56.0         & \multicolumn{1}{c|}{60.7}          & 43.1         & 61.5         & 27.4         & \multicolumn{1}{c|}{44.0}          & 25.8         & 57.6         & 52.9         & \multicolumn{1}{c|}{45.4}          & 50.1                              \\
Qwen2-VL~\cite{wang2024qwen2}                        & \multicolumn{1}{c|}{7B}                               & 69.1         & 53.2         & 63.8         & 50.6         & 66.3         & 60.9         & \multicolumn{1}{c|}{60.7}          & 44.4         & 66.9         & 34.4         & \multicolumn{1}{c|}{48.6}          & 30.1         & 65.7         & 50.8         & \multicolumn{1}{c|}{48.9}          & 52.7                              \\
LLaVA-OneVision~\cite{li2024llava}                 & \multicolumn{1}{c|}{7B}                               & 67.1         & 58.7         & 69.8         & 49.4         & 71.3         & 60.3         & \multicolumn{1}{c|}{62.8}          & 52.5         & 58.8         & 23.7         & \multicolumn{1}{c|}{45.0}          & 24.8         & 66.9         & 60.8         & \multicolumn{1}{c|}{50.9}          & 52.9                              \\
LLaVA-NeXT-Video~\cite{liu2024llavanext}                & \multicolumn{1}{c|}{7B}                               & 69.8         & 59.6         & 66.4         & 50.6         & 72.3         & 61.4         & \multicolumn{1}{c|}{63.3}          & 51.2         & 64.2         & 9.7          & \multicolumn{1}{c|}{41.7}          & 34.1         & 67.6         & 60.8         & \multicolumn{1}{c|}{54.2}          & 53.1                              \\ \hline
\multicolumn{18}{c}{\textbf{Open-Source   Streaming Video-LLMs}}                                                                                                                                                                                                                                                                                                                                                               \\ \hline
VideoLLM-online~\cite{chen2024videollm}                 & \multicolumn{1}{c|}{8B}                               & 8.1          & 23.9         & 12.1         & 14.0         & 45.5         & 21.2         & \multicolumn{1}{c|}{20.8}          & 22.2         & 18.8         & 12.2         & \multicolumn{1}{c|}{17.7}          & -            & -            & -            & \multicolumn{1}{c|}{-}             & -                                 \\
Flash-Vstream~\cite{zhang2025flash}                   & \multicolumn{1}{c|}{7B}                               & 25.5         & 32.1         & 29.3         & 33.7         & 29.7         & 28.8         & \multicolumn{1}{c|}{29.9}          & 36.4         & 33.8         & 5.9          & \multicolumn{1}{c|}{25.4}          & 5.4          & \underline{67.3}         & \underline{60.0}         & \multicolumn{1}{c|}{44.2}          & 33.2                              \\
Dispider~\cite{qian2025dispider}                        & \multicolumn{1}{c|}{7B}                               & 57.7         & 49.5         & 62.1         & 44.9         & 61.4         & 51.6         & \multicolumn{1}{c|}{54.5}          & 48.5         & 55.4         & 4.3          & \multicolumn{1}{c|}{36.1}          & 18.0         & 37.4         & 48.8         & \multicolumn{1}{c|}{34.7}          & 41.8                              \\
TimeChat-Online~\cite{yao2025timechat}                 & \multicolumn{1}{c|}{7B}                               & 69.8         & 48.6         & \underline{64.7}         & 44.9         & \underline{68.3}         & 55.4         & \multicolumn{1}{c|}{58.6}          & 53.9         & \underline{62.8}         & 9.1          & \multicolumn{1}{c|}{\underline{42.0}}          & 32.5         & 36.5         & 40.0         & \multicolumn{1}{c|}{36.4}          & 45.6                              \\
StreamAgent~\cite{yang2025streamagent}                     & \multicolumn{1}{c|}{7B}                               & \underline{71.2}         & \underline{53.2}         & 63.6         & \underline{53.9}         & 67.3         & \underline{58.7}         & \multicolumn{1}{c|}{\underline{61.3}}          & \underline{54.8}         & 58.1         & \underline{25.8}         & \multicolumn{1}{c|}{41.7}          & \underline{35.9}         & 48.4         & 52.0         & \multicolumn{1}{c|}{\underline{45.4}}          & \underline{49.4}                              \\
\textbf{\namewithspace (Ours)}                     & \multicolumn{1}{c|}{8B}                               & \textbf{90.6}         & \textbf{74.3}         & \textbf{75.9}         & \textbf{59.6}         & \textbf{69.3}         & \textbf{71.7}         & \multicolumn{1}{c|}{\textbf{73.6}}          & \textbf{55.6}         & \textbf{66.9}         & \textbf{33.9}         & \multicolumn{1}{c|}{\textbf{52.1}}          & \textbf{41.9}         & \textbf{71.1}         & \textbf{61.7}         & \multicolumn{1}{c|}{\textbf{58.2}}          & \textbf{61.3}                              \\ \hline
\end{tabular}
}
\caption{Performance comparison on OVO-Bench. Following the offical benchmark settings, the Forward Active Responding subtask is implemented using \textit{proactive} interaction mode, where the model must determine when to respond. The Real-Time Visual Perception and Backward Tracing subtask follow the \textit{reactive} interaction mode. Among open‑source streaming Video‑LLMs, the \textbf{best} and \underline{second‑best} scores are highlighted.
}
\label{table:ovobench}
\end{table*}

\begin{table*}[t]
\centering
\resizebox{1.0\linewidth}{!}{
\begin{tabular}{lcccccccccccccc}
\hline
\multirow{2}{*}{\textbf{Model}} & \multicolumn{1}{c|}{\multirow{2}{*}{\textbf{Params}}} & \multicolumn{11}{c|}{\textbf{Real-Time   Visual Understanding}}                                                                                                                 & \multicolumn{1}{c|}{\multirow{2}{*}{\textbf{PO}}} & \multirow{2}{*}{\textbf{\begin{tabular}[c]{@{}c@{}}Overall\\ Avg.\end{tabular}}} \\ \cline{3-13}
                                & \multicolumn{1}{c|}{}                                 & \textbf{OP} & \textbf{CR} & \textbf{CS} & \textbf{ATP} & \textbf{EU} & \textbf{TR} & \textbf{PR} & \textbf{SU} & \textbf{ACP} & \textbf{CT} & \multicolumn{1}{c|}{\textbf{All}} & \multicolumn{1}{c|}{}                             &                                   \\ \hline
\multicolumn{15}{c}{\textbf{Human}}                                                                                                                                                                                                                                                                                                                               \\ \hline
Human                           & \multicolumn{1}{c|}{-}                                & 89.5        & 92.0        & 93.6        & 91.5         & 95.7        & 92.5        & 88.0        & 88.8        & 89.7         & 91.3        & \multicolumn{1}{c|}{91.5}         & \multicolumn{1}{c|}{100}                          & 92.0                              \\ \hline
\multicolumn{15}{c}{\textbf{Proprietary MLLMs}}                                                                                                                                                                                                                                                                                                                   \\ \hline
Claude 3.5 Sonnet~\cite{anthropic2024claude}               & \multicolumn{1}{c|}{-}                                & 80.5        & 77.3        & 82.0        & 81.7         & 72.3        & 75.4        & 61.1        & 61.8        & 69.3         & 43.1        & \multicolumn{1}{c|}{72.4}         & \multicolumn{1}{c|}{64.7}                         & 69.9                              \\
GPT-4o~\cite{hurst2024gpt}                          & \multicolumn{1}{c|}{-}                                & 77.1        & 80.5        & 83.9        & 76.5         & 70.2        & 83.8        & 66.7        & 62.2        & 69.1         & 49.2        & \multicolumn{1}{c|}{73.3}         & \multicolumn{1}{c|}{56.9}                         & 70.5                              \\
Gemini 1.5 pro~\cite{team2024gemini}                  & \multicolumn{1}{c|}{-}                                & 79.0        & 80.5        & 83.5        & 79.7         & 80.0        & 84.7        & 77.8        & 64.2        & 72.0         & 48.7        & \multicolumn{1}{c|}{75.7}         & \multicolumn{1}{c|}{45.1}                         & 72.3                              \\ \hline
\multicolumn{15}{c}{\textbf{Open-Source   Video-LLMs}}                                                                                                                                                                                                                                                                                                            \\ \hline
Video-LLaMa2~\cite{cheng2024videollama}                    & \multicolumn{1}{c|}{7B}                               & 55.9        & 55.5        & 57.4        & 58.2         & 52.8        & 43.6        & 39.8        & 42.7        & 45.6         & 35.2        & \multicolumn{1}{c|}{49.5}         & \multicolumn{1}{c|}{0.0}                          & 44.2                              \\
VILA-1.5~\cite{lin2024vila}                        & \multicolumn{1}{c|}{8B}                               & 53.7        & 49.2        & 71.0        & 56.9         & 43.4        & 53.9        & 54.6        & 48.8        & 50.1         & 17.6        & \multicolumn{1}{c|}{52.3}         & \multicolumn{1}{c|}{17.7}                         & 47.0                              \\
Video-CCAM~\cite{fei2024video}                      & \multicolumn{1}{c|}{14B}                              & 56.4        & 57.8        & 65.3        & 62.8         & 64.6        & 51.4        & 42.6        & 48.0        & 49.6         & 31.6        & \multicolumn{1}{c|}{54.0}         & \multicolumn{1}{c|}{22.7}                         & 50.2                              \\
LongVA~\cite{zhang2024long}                          & \multicolumn{1}{c|}{7B}                               & 70.0        & 63.3        & 61.2        & 70.9         & 62.7        & 59.5        & 61.1        & 53.7        & 54.7         & 34.7        & \multicolumn{1}{c|}{60.0}         & \multicolumn{1}{c|}{15.9}                         & 55.2                              \\
InternVL-V2~\cite{chen2024far}                     & \multicolumn{1}{c|}{8B}                               & 68.1        & 60.9        & 69.4        & 77.1         & 67.7        & 62.9        & 59.3        & 53.3        & 55.0         & 56.5        & \multicolumn{1}{c|}{63.7}         & \multicolumn{1}{c|}{40.9}                         & 61.0                              \\
Kangaroo~\cite{liu2024kangaroo}                        & \multicolumn{1}{c|}{7B}                               & 71.1        & 84.4        & 70.7        & 73.2         & 67.1        & 61.7        & 56.5        & 55.7        & 62.0         & 38.9        & \multicolumn{1}{c|}{64.6}         & \multicolumn{1}{c|}{16.0}                         & 59.7                              \\
LLaVA-NeXT-Video~\cite{liu2024llavanext}                & \multicolumn{1}{c|}{32B}                              & 78.2        & 70.3        & 73.8        & 76.8         & 63.4        & 69.8        & 57.4        & 56.1        & 64.3         & 38.9        & \multicolumn{1}{c|}{67.0}         & \multicolumn{1}{c|}{18.2}                         & 60.6                              \\
MiniCPM-V-2.6~\cite{hu2024minicpm}                   & \multicolumn{1}{c|}{8B}                               & 71.9        & 71.1        & 77.9        & 75.8         & 64.6        & 65.7        & 70.4        & 56.1        & 62.3         & 53.4        & \multicolumn{1}{c|}{67.4}         & \multicolumn{1}{c|}{22.2}                         & 62.9                              \\
LLaVA-OneVision~\cite{li2024llava}                 & \multicolumn{1}{c|}{7B}                               & 80.4        & 74.2        & 76.0        & 80.7         & 72.7        & 71.7        & 67.6        & 65.5        & 65.7         & 45.1        & \multicolumn{1}{c|}{71.1}         & \multicolumn{1}{c|}{29.6}                         & 66.3                              \\
Qwen2-VL~\cite{wang2024qwen2}                        & \multicolumn{1}{c|}{7B}                               & 75.2        & 82.8        & 73.2        & 77.5         & 68.3        & 71.0        & 72.2        & 61.2        & 61.5         & 46.1        & \multicolumn{1}{c|}{69.0}         & \multicolumn{1}{c|}{22.7}                         & 64.7                              \\ \hline
\multicolumn{15}{c}{\textbf{Open-Source   Streaming Video-LLMs}}                                                                                                                                                                                                                                                                                                  \\ \hline
VideoLLM-online~\cite{chen2024videollm}                 & \multicolumn{1}{c|}{8B}                               & 39.1        & 40.1        & 34.5        & 31.1         & 46.0        & 32.4        & 31.5        & 34.2        & 42.5         & 27.9        & \multicolumn{1}{c|}{36.0}         & \multicolumn{1}{c|}{3.9}                          & 33.0                              \\
Flash-Vstream~\cite{zhang2025flash}                   & \multicolumn{1}{c|}{7B}                               & 25.9        & 43.6        & 24.9        & 23.9         & 27.3        & 13.1        & 18.5        & 25.2        & 23.9         & 48.7        & \multicolumn{1}{c|}{23.2}         & \multicolumn{1}{c|}{2.0}                          & 25.2                              \\
Dispider~\cite{qian2025dispider}                        & \multicolumn{1}{c|}{7B}                               & 74.9        & 75.5        & 74.1        & 73.1         & 74.4        & 59.9        & 76.1        & 62.9        & 62.2         & 45.8        & \multicolumn{1}{c|}{67.6}         & \multicolumn{1}{c|}{25.3}                         & 64.0                              \\
TimeChat-Online~\cite{yao2025timechat}                 & \multicolumn{1}{c|}{7B}                               & \underline{80.2}        & \textbf{82.0}        & \underline{79.5}        & \underline{83.3}         & \textbf{76.1}        & \underline{78.5}        & 78.7        & \underline{64.6}        & \underline{69.6}         & \textbf{58.0}        & \multicolumn{1}{c|}{\underline{75.4}}         & \multicolumn{1}{c|}{28.8}                         & \underline{70.9}                              \\
StreamAgent~\cite{yang2025streamagent}                     & \multicolumn{1}{c|}{7B}                               & 79.6        & \underline{78.3}        & 79.3        & 75.9         & 74.7        & 76.9        & \underline{82.9}        & \textbf{66.3}        & \textbf{73.7}         & 55.4        & \multicolumn{1}{c|}{74.3}         & \multicolumn{1}{c|}{\underline{28.9}}                         & 70.2                              \\
\textbf{\namewithspace(Ours)}                     & \multicolumn{1}{c|}{8B}                               & \textbf{84.0}        & 78.1        & \textbf{88.0}        & \textbf{84.6}         & \underline{74.8}        & \textbf{83.5}        & \textbf{83.3}        & 63.0        & 69.3         & \textbf{58.0}        & \multicolumn{1}{c|}{\textbf{77.5}}         & \multicolumn{1}{c|}{\textbf{44.0}}                         & \textbf{73.7}                              \\ \hline
\end{tabular}
}
\caption{Performance comparison on StreamingBench.
Following the official benchmark settings, the PO (Proactive Output) subtask implemented using \textit{proactive} interaction mode.
All other subtasks follow the \textit{reactive} interaction mode. Among open‑source streaming Video‑LLMs, the \textbf{best} and \underline{second‑best} scores are highlighted.}
\label{table:streamingbench}
\end{table*}

%% file: sec/4_experiment.tex
\begin{table}[t]
\centering
\resizebox{1\linewidth}{!}{
\begin{tabular}{l|ccc|ccc}
\hline
\multirow{2}{*}{\textbf{Strategies}} & \multicolumn{3}{c|}{\textbf{OVO-Bench}}     & \multicolumn{3}{c}{\textbf{StreamingBench}}            \\ \cline{2-7} 
                            & \textbf{ACR}   & \multicolumn{1}{c|}{\textbf{HLD}}   & \textbf{CRR}   & \textbf{CS}  & \multicolumn{1}{c|}{\textbf{PR}}  & \textbf{PO}  \\ \hline
W/o Retrieval Augmentation            & 66.1 & \multicolumn{1}{c|}{28.0} & 55.4 & 83.6 & \multicolumn{1}{c|}{79.6} & 36.8 \\
W/o Timestamp Encoding              & 74.0 & \multicolumn{1}{c|}{33.6} & 60.4 & 87.7 & \multicolumn{1}{c|}{82.9} & 43.6 \\
Full                        & \textbf{74.3} & \multicolumn{1}{c|}{\textbf{33.9}} & \textbf{61.7} & \textbf{88.0} & \multicolumn{1}{c|}{\textbf{83.3}} & \textbf{44.0} \\ \hline
\end{tabular}
}
\caption{Performance of different retrieval-augmented streaming inference strategies. Evaluations are conducted on \textit{proactive} subtasks (CRR, PO) and \textit{reactive} subtasks (ACR, HLD, CS, PR).}
\label{table:ab_svu}
\end{table}

\section{Experiments}
\label{sec:experiment}

\subsection{Settings}
\paragraph{Datasets.}
We evaluate \namewithspace on two established streaming video understanding benchmarks: OVO-Bench~\cite{niu2025ovo} and StreamingBench~\cite{lin2024streamingbench}.
The evaluations are conducted in two distinct modes:
(i) \textbf{Proactive mode}, which aims to validate \namewithspace’s capability for more accurate response timing decisions. This mode employs the Forward Active Responding subtask in OVO‑Bench and the PO (Proactive Output) subtask in StreamingBench, where the model must autonomously decide when to respond during the video stream.
(ii) \textbf{Reactive mode}, which verifies the versatility of our framework by demonstrating improved spatio-temporal understanding through scene-graph-retrieved augmentation. This mode includes all remaining subtasks of the two benchmarks, where the response timing is pre‑specified and coincides with the moment the user query is issued.


\paragraph{Implementation Details.}
We employ Qwen3-VL-8B~\cite{bai2025qwen3vltechnicalreport} as our Video-LLM backbone, and follow the input pixel configuration established in prior work~\cite{yao2025timechat}.
For OVO-Bench~\cite{niu2025ovo}, evaluations are performed at default 1 FPS.
For StreamingBench~\cite{lin2024streamingbench}, we follow its official frame‑sampling protocol: videos shorter than 300 frames are sampled at 1 FPS, those between 300 and 600 frames at 0.5 FPS, and videos longer than 600 frames at 0.2 FPS.
All experiments are run on NVIDIA A100 (80GB) GPUs using FP16 precision.


\subsection{Main Results}
\paragraph{Proactive Streaming Video Understanding.}
For the \textit{proactive} evaluation, we report results on the Forward Active Responding subtask in OVO-Bench, namely REC (Repetition Event Count), SSR (Sequential Steps Recognition), and CRR (Clues Reveal Responding), as summarized in Table~\ref{table:ovobench}.
The results demonstrate our leading performance across these subtasks. Notably, the average score over the three subtasks surpasses the second-best open‑source streaming Video‑LLM by 12.8\%, and even attains a level comparable to proprietary MLLMs.
For StreamingBench, the performance on the PO (Proactive Output) subtask is presented in Table~\ref{table:streamingbench}.
\namewithspace outperforms the second-best open-source streaming Video‑LLM by 15.1\%.
These results further highlight the superiority of \namewithspace in proactive streaming video understanding tasks.

\paragraph{Reactive Streaming Video Understanding.}
We also evaluate \namewithspace on standard \textit{reactive} streaming video understanding tasks.
As reported in Table~\ref{table:ovobench}, \namewithspace outperforms the second‑best open‑source streaming Video‑LLM by 12.3\% on the Real‑Time Visual Perception subtask and by 10.1\% on the Backward Tracing subtask in OVO‑Bench.
In StreamingBench shown in Table~\ref{table:streamingbench}, it achieves a 2.1\% improvement on the Real‑Time Visual Understanding subtask.
The consistent performance gains indicate that explicit scene graph modeling and retrieval not only lead to more accurate response timing in \textit{proactive} settings, but also deliver substantial benefits for spatio-temporal understanding of objects and their relations on \textit{reactive} subtasks.


\begin{table}[t]
\centering
\resizebox{0.75\linewidth}{!}{
\begin{tabular}{l|cccc}
\hline
\multirow{2}{*}{\textbf{Strategies}} & \multicolumn{4}{c}{\textbf{Proactive Subtasks}} \\ \cline{2-5} 
                                     & \textbf{PO}  & \textbf{REC}  & \textbf{SSR}  & \textbf{CRR} \\ \hline
W/o Guidance                         & 38.8         & 34.1          & 66.9          & 59.4         \\
Object-Guidance                       & 43.6         & 40.2          & 67.9          & 61.3         \\
Query-Guidance                       & \textbf{44.0}         & \textbf{41.9}          & \textbf{71.1}          & \textbf{61.7}         \\ \hline
\end{tabular}
}
\caption{Performance of different guidance strategies for online video scene graph generation. Evaluations are conducted on \textit{proactive} subtasks.}
\label{table:ab_sgg}
\end{table}

\begin{figure}[t]
\includegraphics[width=0.9\columnwidth]{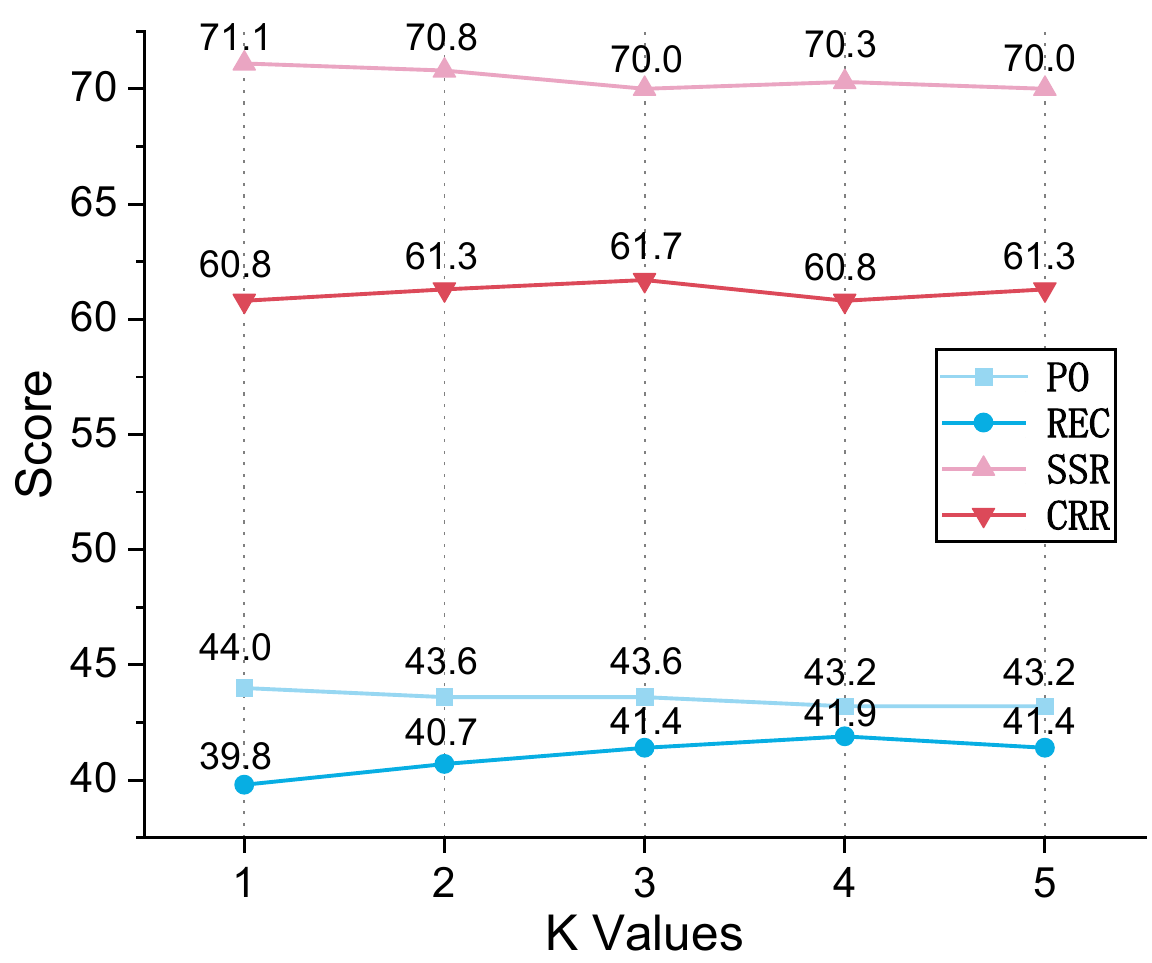}
\caption{Performance of different K Values for top-K similarity-based scene graph retrieval. Evaluations are conducted on \textit{proactive} subtasks.}
\vspace{-5pt}
\label{fig:topk}
\end{figure}

\begin{table}[t]
\centering
\resizebox{0.8\linewidth}{!}{
\begin{tabular}{l|cccc}
\hline
\multirow{2}{*}{\textbf{Strategies}} & \multicolumn{4}{c}{\textbf{Proactive Subtasks}} \\ \cline{2-5} 
                                     & \textbf{PO}  & \textbf{REC}  & \textbf{SSR}  & \textbf{CRR} \\ \hline
Original Query Text                      & 42.4         & 40.2          & 69.3          & 56.0         \\
Query Graph Text                & \textbf{44.0}         & \textbf{41.9}          & \textbf{71.1}          & \textbf{61.7}         \\ \hline
\end{tabular}
}
\caption{Performance of different query embedding strategies for similarity-based scene graph retrieval. Evaluations are conducted on \textit{proactive} subtasks.}
\label{table:ab_similarity}
\end{table}

\begin{figure*}[t]
\centering
  \includegraphics[width=1\textwidth]{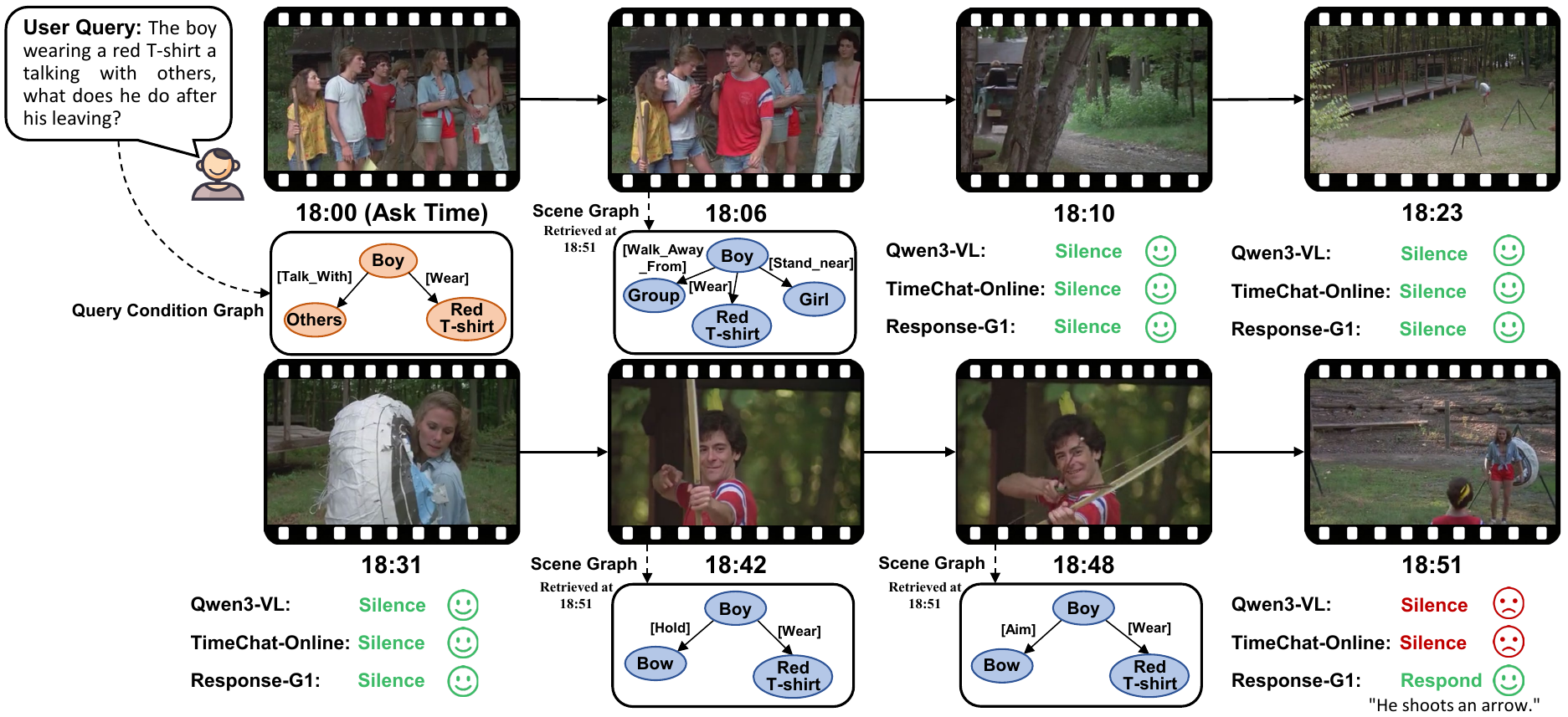}
  \caption{Case study of \namewithspace on the CRR subtask in OVO-Bench. The user query describes a target object (“the boy wearing a red T‑shirt”) and a relation (“talking with others”). The results show that at time “18:51”, \namewithspace accurately retrieves query‑relevant scene graphs (i.e., evidence) and triggers a response, whereas the baselines fail to respond throughout the video stream.
  }
\label{fig:case}
\end{figure*}

\subsection{Ablation Study}
\label{subsec:ablation}
\paragraph{Impact of Retrieval-Augmented Streaming Pipeline.}
To verify whether explicit scene‑graph modeling enhances the Video‑LLM’s streaming video understanding, we compare performance with and without graph‑based retrieval augmentation.
Additionally, we examine the impact of introducing timestamp encoding (refer to §\ref{subsec:pipeline}).
As shown in Table~\ref{table:ab_svu}, our explicit scene‑graph modeling improves performance on both \textit{reactive} and \textit{proactive} subtasks.
Moreover, timestamp‑aware scene‑graph encoding yields a notable gain on tasks that require temporal grounding, such as CRR.

\paragraph{Configurations of Online Query-Guided Scene Graph Generation.}
As outlined in §\ref{subsec:onlinesgg}, we inject the user query into the prompts for online video scene graph generation to guide the Video-LLM toward query‑relevant descriptions and reduce redundant triplet generation (i.e., "Query-Guidance").
The effectiveness of this design is validated through performance comparisons on \textit{proactive} subtasks. We also explore an alternative guidance strategy that directly injects parsed objects and relations into the prompts for scene graph generation (i.e., "Object-Guidance").
We provide the specific prompts and cases in the Appendix.
As shown in Table~\ref{table:ab_sgg}, the query‑guided generation strategy yields the best performance, confirming its efficacy. Failure cases in "Object-Guidance" reveal that direct object injection may lead to hallucination, where the model over‑focuses on anticipated objects and generates non‑existent triplets, causing premature responses and performance drops. This underscores the importance of balancing query relevance with factuality in LLM-based SGG.


\paragraph{Effectiveness of Memory-Based Scene Graph Retrieval.}
We evaluate the effectiveness of the proposed memory-based scene graph retrieval described in §\ref{subsec:memorysgr}.
Specifically, we compare the performance of using original query text versus the proposed graph text for embedding similarity calculation on \textit{proactive} subtasks, as shown in Table~\ref{table:ab_similarity}.
The results indicate that using the original query text for embedding similarity leads to format inconsistency between the query and the video scene graph, which consequently degrades both retrieval quality and downstream task performance. We provide the specific cases in the Appendix.

We further analyze the impact of the top‑K parameter in \textit{proactive} settings, as shown in Figure~\ref{fig:topk}.
The results reveal that overall performance remains relatively stable across different $K$ values.
Moreover, for tasks that primarily focus on latest‑frame information (e.g., SSR and PO), $K=1$ is sufficient to achieve superior performance.

\subsection{Case Study}
We conduct a case study to visualize our explicit scene graph modeling and graph‑retrieval-augmented proactive streaming pipeline.
For comparison, we select two advanced open‑source Video‑LLMs: Qwen3‑VL~\cite{yang2025qwen3} and TimeChat‑Online~\cite{yao2025timechat}.
The case is drawn from the CRR (Clues Reveal Responding) subtask in OVO‑Bench. During the stream, each model is evaluated on whether it correctly remains silent or responds at the appropriate time.
As illustrated in Figure~\ref{fig:case}, \namewithspace achieves better understanding and decision-making of whether the accumulated evidence satisfies the query‑specific response conditions through explicit scene graph modeling and retrieval.
Moreover, it illustrates how our method provides more interpretable evidence retrieval for response timing decisions.




%% file: sec/5_conclusion.tex
\section{Conclusion}
\label{sec:conclusion}
We introduce \name, a scene‑graph‑driven pipeline for proactive streaming video understanding.
By modeling both visual evidence and query-specific response conditions in an explicit, structured scene graph representation, our method enables Video‑LLMs to achieve better understanding of evidence-condition alignment and more accurate response timing decisions, within a fine‑tuning‑free manner.
Superior results across established benchmarks illustrate that explicit structural modeling can be a powerful principle for proactive interaction.
We hope this work opens promising avenues toward more capable and interpretable multimodal interaction in real-world streaming applications.

%% file: sec/appendix.tex
\section*{Appendix}
\label{sec:appendix}

\section*{A \quad Latency Analysis}
To assess real-world feasibility, we implement latency analysis of \namewithspace and a naive Qwen3-VL-8B baseline on the PO subtask in StreamingBench (1 FPS sampling). Latency is defined as the average per-frame execution time from frame input to scene graph generation (if applicable), scene graph retrieval (if applicable), and trigger decision upon each frame arrival. Maximum FPS, computed as $1s / \text{Total Latency}$, represents the maximum sampling rate supported by the given configuration.

Additionally, we incorporate a streaming KV-Cache mechanism: instead of storing historical frames as visual embeddings, the memory bank stores them as KV caches, avoiding redundant computation during streaming inference.

As shown in Table~\ref{table:latency}, \namewithspace achieves a maximum FPS of 1.2, which satisfies the 1 FPS sampling rate. With the streaming KV-Cache mechanism, the maximum FPS further increases to 2.1 without sacrificing accuracy, confirming the real-world feasibility of our method.

\begin{table}[h]
\centering
\resizebox{1\linewidth}{!}{
\begin{tabular}{l|c|cccc|c}
\hline
\textbf{Model}               & \textbf{Memory} & \textbf{\begin{tabular}[c]{@{}c@{}}SGG\\ (ms)\end{tabular}} & \textbf{\begin{tabular}[c]{@{}c@{}}SGR\\ (ms)\end{tabular}} & \textbf{\begin{tabular}[c]{@{}c@{}}Trigger\\ (ms)\end{tabular}} & \textbf{\begin{tabular}[c]{@{}c@{}}Total\\ (ms)\end{tabular}} & \textbf{\begin{tabular}[c]{@{}c@{}}Maximum\\ FPS\end{tabular}} \\ \hline
\multirow{2}{*}{Qwen3-VL-8B} & Embedding       & \textbackslash{}                                            & \textbackslash{}                                            & 324                                                             & 324                                                           & 3.1                                                            \\
                             & KV-Cache        & \textbackslash{}                                            & \textbackslash{}                                            & 182                                                             & 182                                                           & 5.5                                                            \\ \hline
\multirow{2}{*}{\name} & Embedding       & 448                                                         & 21                                                          & 356                                                             & 825                                                           & 1.2                                                            \\
                             & KV-Cache        & 249                                                         & 20                                                          & 204                                                             & 473                                                           & 2.1                                                            \\ \hline
\end{tabular}
}
\caption{Latency analysis on the PO subtask. SGG: Scene Graph Generation; SGR: Scene Graph Retrieval.}
\label{table:latency}
\end{table}

\section*{B \quad Evaluation Across Architectures}
We extend our evaluation to LLaVA-OneVision-1.5-8B~\cite{an2025llava}, the latest open-source model in the LLaVA-OneVision series.
As shown in Table~\ref{table:backbone_comparison}, which presents a comparison among open-source streaming Video-LLMs, \namewithspace(i.e., Ours) consistently achieves leading performance on both benchmarks when deployed across different architectures. These results demonstrate the generalizability of our scene-graph-driven approach across diverse Video-LLM architectures.

\begin{table}[h]
\centering
\resizebox{1\linewidth}{!}{
\begin{tabular}{ll|cccc|ccc}
\hline
\multicolumn{2}{l|}{\multirow{2}{*}{\textbf{Model}}}             & \multicolumn{4}{c|}{\textbf{OVO-Bench}}                       & \multicolumn{3}{c}{\textbf{StreamingBench}}    \\ \cline{3-9} 
\multicolumn{2}{l|}{}                                            & \textbf{RTVP} & \textbf{BT} & \textbf{FAR} & \textbf{Overall} & \textbf{RTVU} & \textbf{PO} & \textbf{Overall} \\ \hline
\multicolumn{2}{l|}{VideoLLM-online}                    & 20.8          & 17.7        & -            & -                & 36.0          & 3.9         & 33.0             \\
\multicolumn{2}{l|}{Flash-Vstream}                      & 29.9          & 25.4        & 44.2         & 33.2             & 23.2          & 2.0         & 25.2             \\
\multicolumn{2}{l|}{Dispider}                           & 54.5          & 36.1        & 34.7         & 41.8             & 67.6          & 25.3        & 64.0             \\
\multicolumn{2}{l|}{TimeChat-Online}                    & 58.6          & 42.0        & 36.4         & 45.6             & \underline{75.4}          & 28.8        & 70.9             \\
\multicolumn{2}{l|}{StreamAgent}                        & 61.3          & 41.7        & 45.4         & 49.4             & 74.3          & 28.9        & 70.2             \\ \hline
\multirow{2}{*}{\textbf{Ours}} & \textbf{LLaVA-OV-1.5-8B} & \underline{66.7}          & \underline{48.3}        & \underline{54.8}         & \underline{56.1}             & 74.8          & \underline{35.6}        & \underline{71.2}             \\
                                      & \textbf{Qwen3-VL-8B}     & \textbf{73.6}          & \textbf{52.1}        & \textbf{58.2}         & \textbf{61.3}             & \textbf{77.5}          & \textbf{44.0}        & \textbf{73.7}             \\ \hline
\end{tabular}
}
\caption{Performance comparison among open-source streaming Video-LLMs on OVO-Bench and StreamingBench. RTVP: Real-Time Visual Perception; BT: Backward Tracing; FAR: Forward Active Responding; RTVU: Real-Time Visual Understanding; PO: Proactive Output.}
\label{table:backbone_comparison}
\end{table}

\section*{C \quad Prompts}
\label{sec:prompts}
This section provides the detailed prompts used in \namewithspace for (i) scene graph generation (video clip → textual scene graph), (ii) query parsing (user query → textual query condition graph), and (iii) graph-retrieval-augmented streaming trigger.

\begin{figure*}[h]
\centering
  \includegraphics[width=1\textwidth]{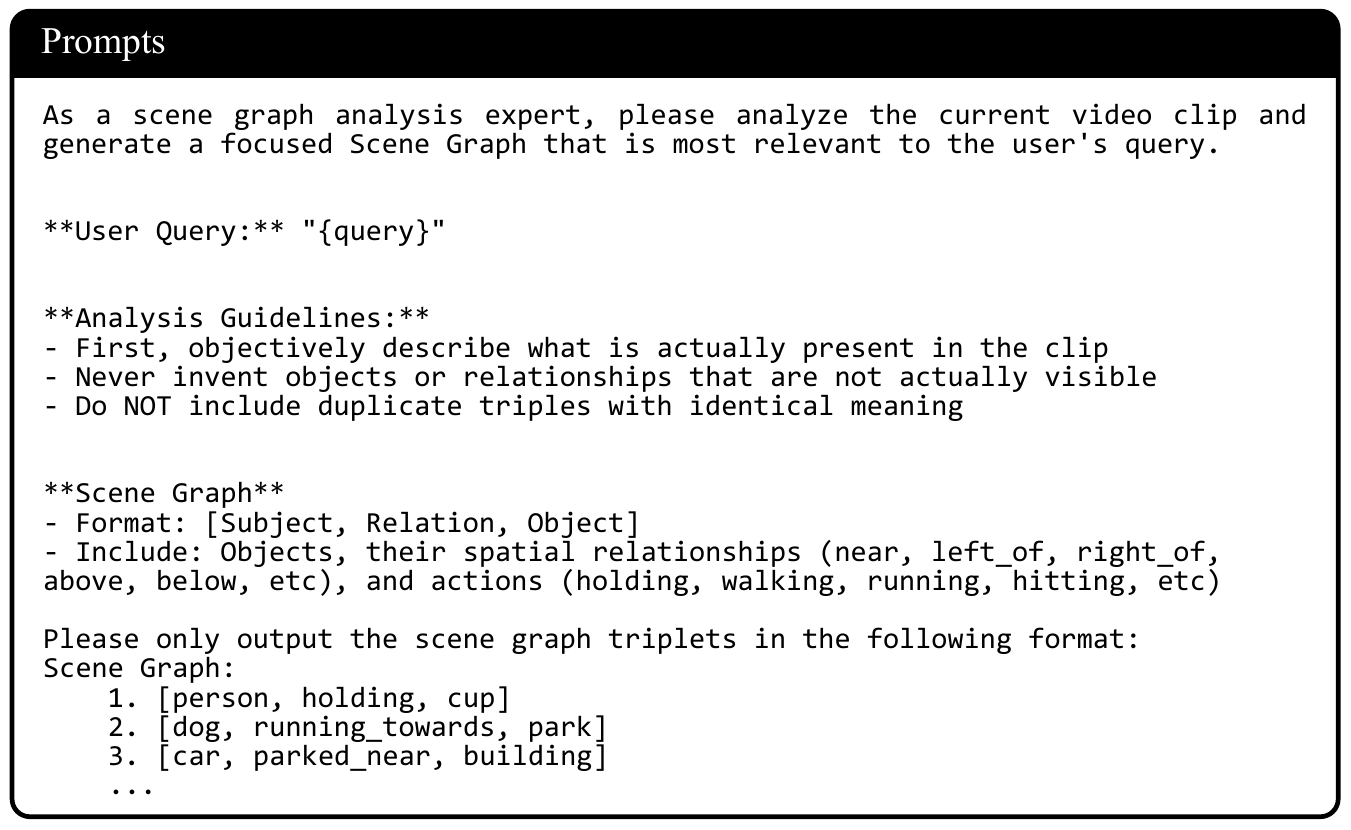}
  \caption{Prompt template for query-guided online scene graph generation.}
  \label{fig:sgg_prompt}
\end{figure*}

\paragraph{Scene Graph Generation Prompts}
As described in §\ref{subsec:onlinesgg}, we perform query-guided scene graph generation on streaming video clips. Figure~\ref{fig:sgg_prompt} illustrates the complete prompt template used for this process.

In Table~\ref{table:ab_sgg}, we validate three guidance strategies for scene graph generation. Below, we illustrate each using a concrete case from the PO subtask in StreamingBench, where the query asks to respond when \texttt{"the number 20 appears in the middle of the sun"}.
\begin{itemize}
    \item "W/o Guidance" relies solely on video clips (\texttt{\{query\} None}). Lacking query awareness, it may generate irrelevant triplets (e.g., \texttt{[grass, on, ground]}), introducing noise for retrieval.
    \item "Object-Guidance" takes parsed query elements as input (\texttt{\{query\} objects: Sun, 20. relations: appear\_in}). Over-focusing on given elements may cause hallucination (e.g., generating the scene graph triplet \texttt{[number 20, appears\_in, sun]} before evidence actually appears), leading to premature triggering.
    \item "Query-Guidance" uses the original query (\texttt{\{query\} When the number 20 appears in the middle of the sun in the video, output "20"}). Experimental results show that it maintains query relevance in scene graph generation while avoiding over-commitment to objects not yet visible, leading to more accurate response timing decisions.
\end{itemize}

\begin{figure*}[h]
\centering
  \includegraphics[width=1\textwidth]{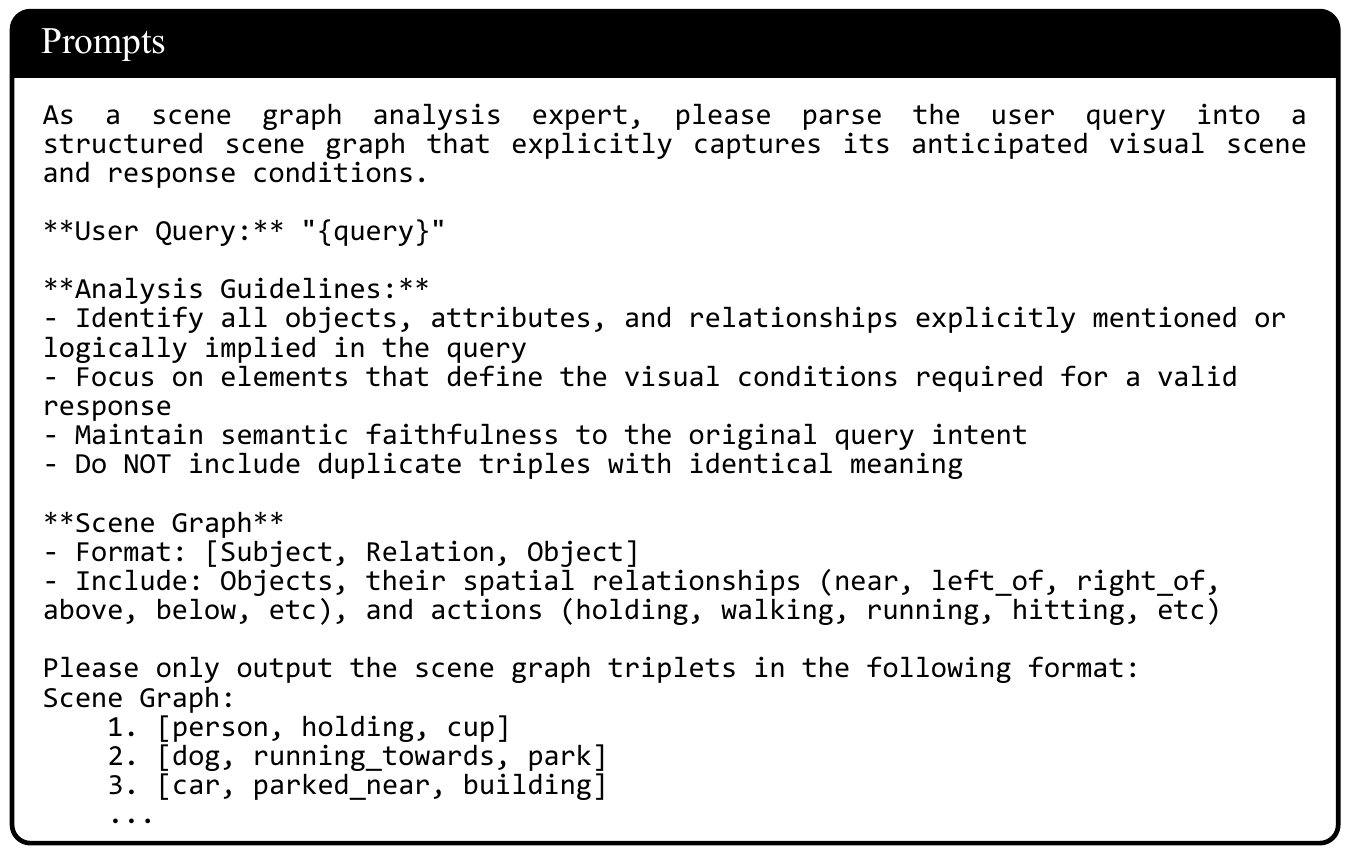}
  \caption{Prompt template for query parsing.}
  \label{fig:qp_prompt}
\end{figure*}

\paragraph{Query Parsing Prompts}
As described in §\ref{subsec:memorysgr}, we perform scene graph retrieval to enable explicit evidence-condition alignment for subsequent trigger decisions.
However, as shown in Table~\ref{table:ab_similarity}, directly using the original query text (e.g., \texttt{"the number 20 appears in the middle of the sun"}) leads to format inconsistency between the video scene graph and the user query, which degrades retrieval and downstream task performance.
To address this, we also parse the user query into a unified structured graph representation (e.g., \texttt{[number 20, appears\_in, sun]}) immediately upon issuance, thereby eliminating the format inconsistency that hinders similarity-based retrieval.
Figure~\ref{fig:qp_prompt} provides the complete prompt template for this query parsing process.

\begin{figure*}[h]
\centering
  \includegraphics[width=1\textwidth]{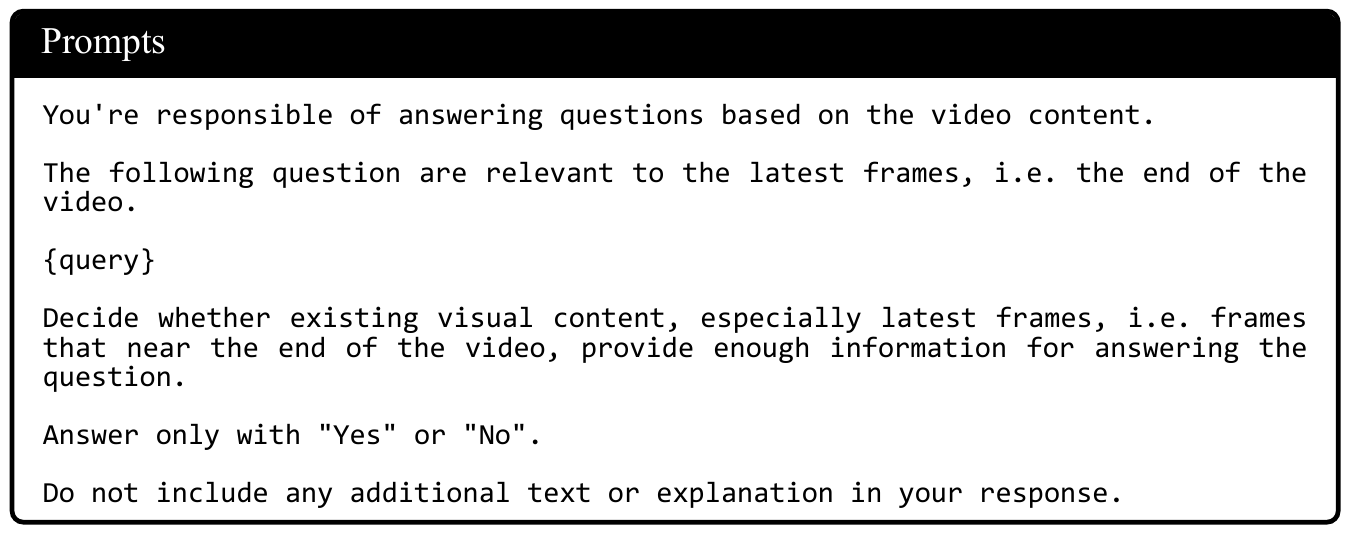}
  \caption{Prompt template for the original trigger on the CRR subtask in OVO-Bench.}
  \label{fig:crr_original_prompt}
\end{figure*}

\begin{figure*}[h]
\centering
  \includegraphics[width=1\textwidth]{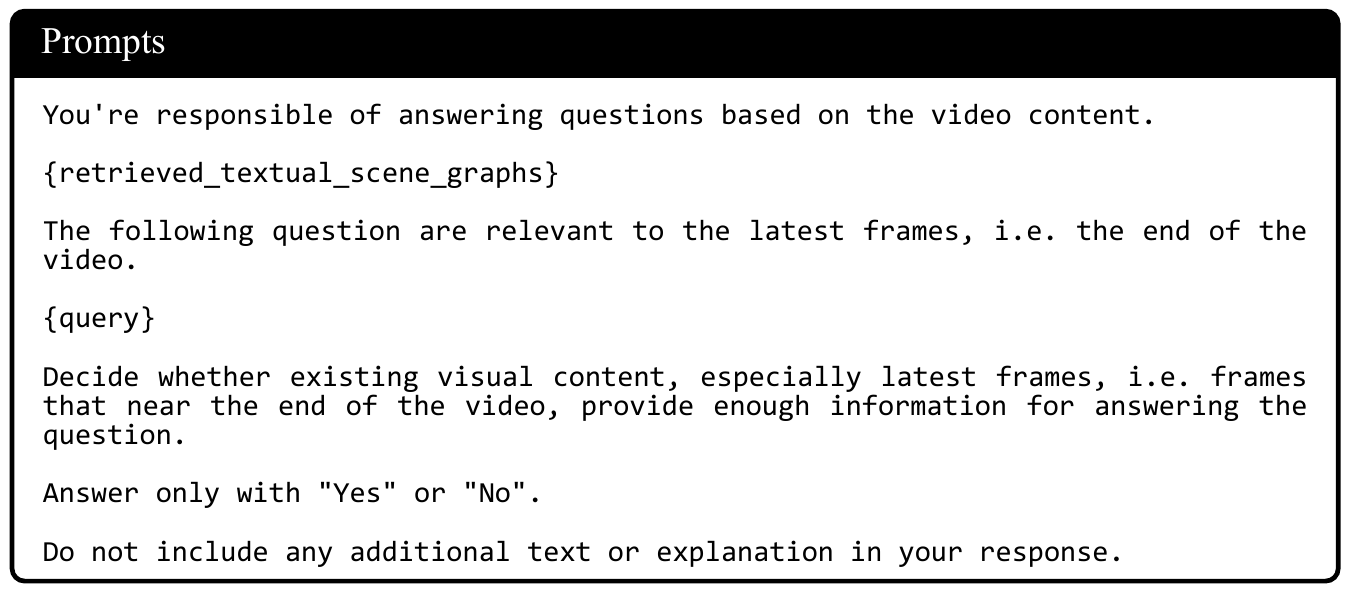}
  \caption{Prompt template for \name 's trigger on the CRR subtask in OVO-Bench.}
  \label{fig:crr_ours_prompt}
\end{figure*}

\begin{figure*}[h]
\centering
  \includegraphics[width=1\textwidth]{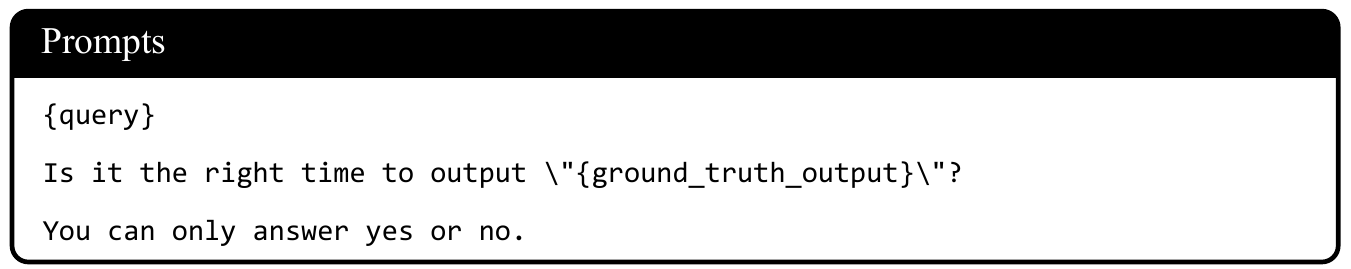}
  \caption{Prompt template for the original trigger on the PO subtask in StreamingBench.}
  \label{fig:po_original_prompt}
\end{figure*}

\begin{figure*}[h]
\centering
  \includegraphics[width=1\textwidth]{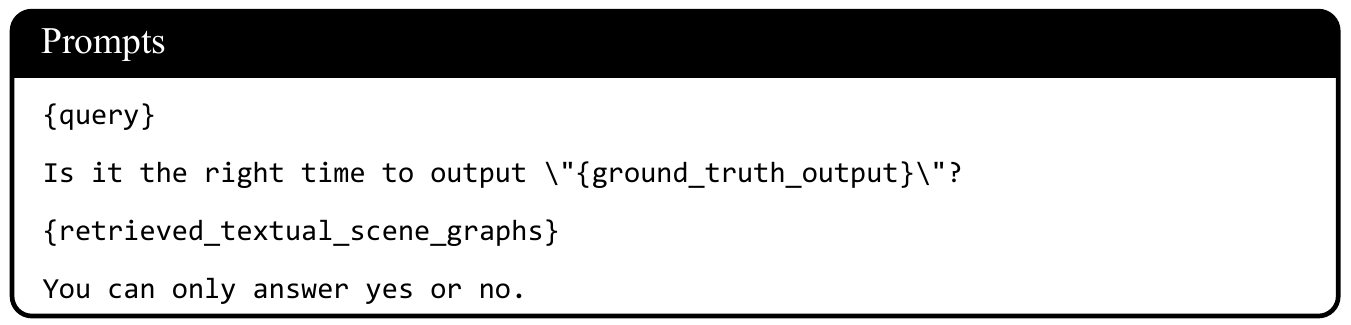}
  \caption{Prompt template for \name 's trigger on the PO subtask in StreamingBench.}
  \label{fig:po_ours_prompt}
\end{figure*}

\paragraph{Retrieval-Augmented Trigger Prompts}
As described in §\ref{subsec:pipeline}, \namewithspace performs retrieval-augmented trigger decisions. Figures~\ref{fig:crr_original_prompt}, \ref{fig:crr_ours_prompt}, \ref{fig:po_original_prompt}, and \ref{fig:po_ours_prompt} show the trigger decision prompt templates for the CRR subtask in OVO-Bench (original and ours) and the PO subtask in StreamingBench (original and ours), respectively.

